\documentclass[11pt,journal]{IEEEtran}

\usepackage{times}
\usepackage{epsfig}
\usepackage{graphicx}
\usepackage{amsmath}
\usepackage{amssymb}
\usepackage{hhline}
\usepackage{booktabs}
\usepackage{cite}
\usepackage{epstopdf}
\usepackage{subcaption}
\usepackage{caption}
\usepackage{sidecap}
\usepackage{url}
\usepackage{color}
\usepackage[noend]{algpseudocode}
\usepackage{algorithm}
\usepackage{mathtools}

\def\imagebox#1#2{\vtop to #1{\null\hbox{#2}\vfill}}

\DeclarePairedDelimiter\ceil{\lceil}{\rceil}
\DeclarePairedDelimiter\floor{\lfloor}{\rfloor}

\begin{document}
\title{A Spatial Layout and Scale Invariant Feature Representation for Indoor Scene Classification}

%
%
%
%
%

\author{M.~Hayat, 
        S.~H.~Khan, 
        M. Bennamoun, 
        and~S.~An, 

\thanks{ M.~Hayat, S.~H.~Khan, M.~Bennamoun and S.~An are with the School of Computer Science and Software Engineering (CSSE), The University of Western Australia, Crawley, 6009. E-mail: munawar.hayat@research.uwa.edu.au, \{salman.khan, mohammed.bennamoun, senjian.an\}@uwa.edu.au}
\thanks{}} 


\maketitle

\begin{abstract}
Unlike standard object classification, where the image to be classified contains one or multiple instances of the same object, indoor scene classification is quite different since the image consists of multiple distinct objects. Further, these objects can be of varying sizes and are present across numerous spatial locations in different layouts. For automatic indoor scene categorization,
large scale spatial layout deformations and scale variations are therefore two major challenges and the design of rich feature descriptors which are robust to these challenges is still an open problem.
This paper introduces a new learnable feature descriptor called ``spatial layout and scale invariant convolutional activations" to deal with these challenges. For this purpose, a new Convolutional Neural Network architecture is designed which incorporates a novel `Spatially Unstructured' layer to introduce robustness against spatial layout deformations.
To achieve scale invariance, we present a pyramidal image representation.
For feasible training of the proposed network for images of indoor scenes, the paper proposes a new methodology which efficiently adapts a trained network model (on a large scale data) for our task with only a limited amount of available training data.
Compared with existing state of the art, the proposed approach achieves a relative performance improvement of 3.2\%, 3.8\%, 7.0\%, 11.9\% and 2.1\% on MIT-67, Scene-15, Sports-8, Graz-02 and NYU datasets respectively.
\end{abstract}

 \maketitle
\section{Introduction}

{\color{black}Recognition/classification is an important computer vision problem and has gained significant research attention over last few decades. Most of the efforts, in this regard, has been tailored towards generic object recognition (an image with one or multiple instances of the same object) and face recognition (an image with the face region of the person). Unlike these classification tasks, indoor scene classification is quite different since an image of an indoor scene contains multiple distinct objects, with different scales and sizes and laid across different spatial locations in a number of possible layouts. Due to the challenging nature of the problem, the state of the art performance for indoor scene classification is much lower (69\% classification accuracy on MIT-67 dataset with only 67 classes \cite{GongMOP14}) compared with other classification tasks such as object classification (94\% rank-5 identification rate on ImageNet database with 1000 object categories \cite{simonyan2014very}) and face recognition (human level performance on face recognition on real life datasets including Labeled Faces in the Wild and YouTube Faces \cite{taigman2014deepface}). This paper proposes a novel method of feature description, specifically tailored for indoor scene images, in order to address the challenges of large scale spatial layout deformations and scale variations.}

\begin{figure}[t!]
\centering
\includegraphics[clip, trim=0 0 0 0 , width=1\columnwidth]{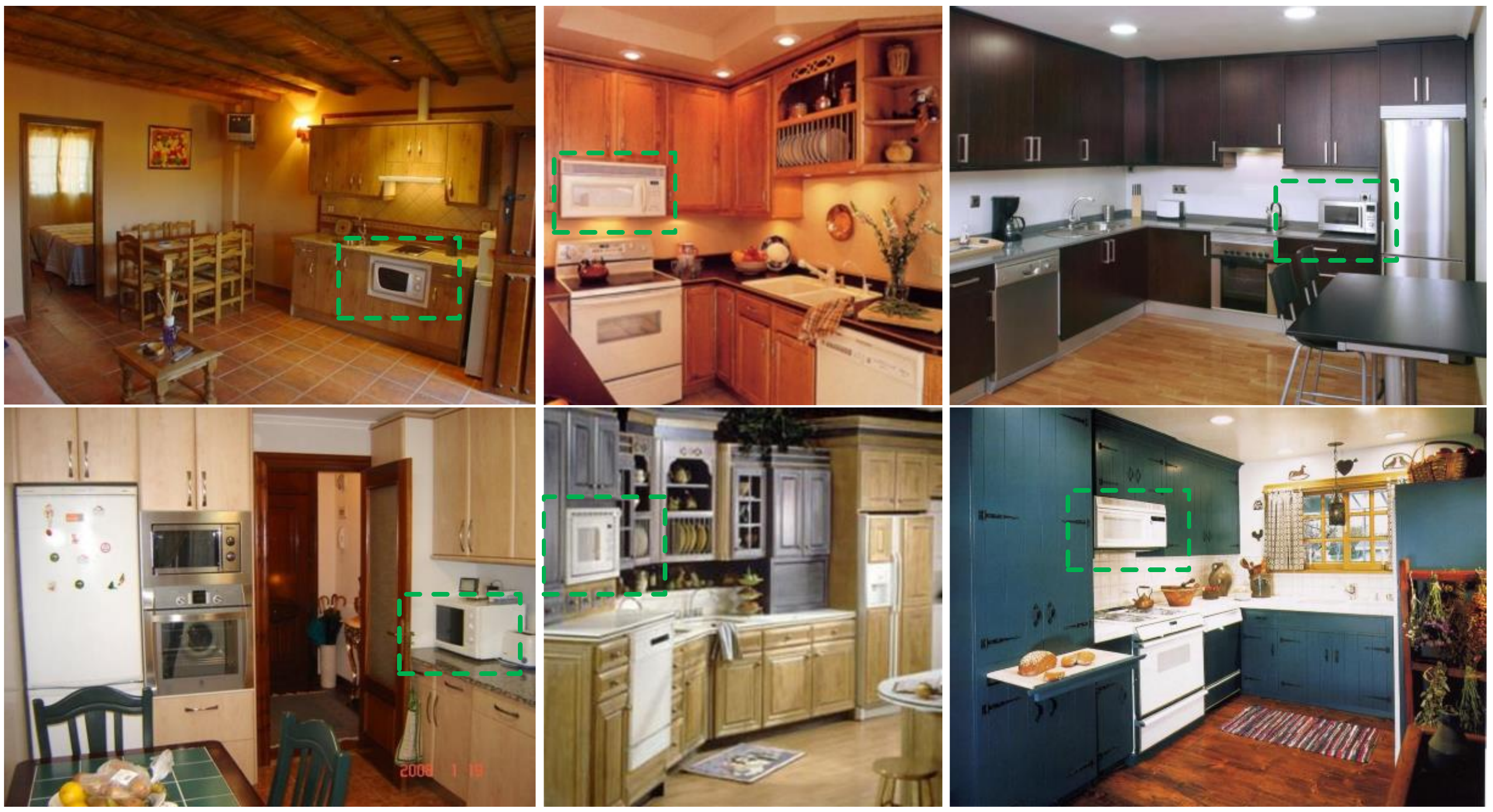}
\caption{The spatial structure of indoor scenes is loose, irregular and unpredictable which can confuse the classification system.
As an example, a microwave in a kitchen scene can be close to the sink, fridge, kitchen door or top cupboards (\emph{green} box in the images). Our objective is to learn feature representations which are robust to these variations by spatially shuffling the convolutional activations (Sec.~\ref{Proposed Spatial Layout and Scale Invariant Convolutional Activations}).}
\label{fig:SceneExample}
\end{figure}

We can characterize some indoor scenes by only global spatial information \cite{oliva2001modeling, razavian2014cnn}, whereas for others, local appearance information \cite{fei2005bayesian, lazebnik2006beyond,  margolin2014otc} is more critical.
For example, a corridor can be predominantly characterized by a single large object (\emph{walls}) whereas a bedroom scene is characterized by multiple objects (e.g, \emph{sofa}, \emph{bed}, \emph{table}).
Both global and local spatial information must therefore be leveraged in order to accommodate different scene types \cite{quattoni2009recognizing}.
This however is very challenging, for two main reasons. \textbf{First}, the spatial scale of the constituent objects varies significantly across different scene types.
\textbf{Second}, the constituent objects can be present in different spatial locations and in a number of possible layouts.
This is demonstrated in the example images of the kitchen scene in Fig. \ref{fig:SceneExample}, where a microwave can be present in many different locations in the image with significant variations in scale, pose and appearance.


This paper aims to achieve invariance with respect to the spatial layout and the scale of the constituent objects for indoor scene images. For this purpose, in order to achieve invariance with respect to the spatial scale of objects, we generate a pyramidal image representation where an image is resized to different scales, and features are computed across these scales (Sec~\ref{Pyramid Image Representation}).
To achieve spatial layout invariance, we introduce a new method of feature description which is based on a proposed modified Convolutional Neural Network (CNN) architecture (Sec.~\ref{subsec:CNNArch}).

CNNs preserve the global spatial layout in an image.
This is desirable for the classification tasks where an image predominantly contains only a single object (e.g., objects in ImageNet database \cite{ILSVRCarxiv14}).
However, for a high level vision task such as indoor scene classification, an image may contain multiple distinct objects across different spatial locations.
We therefore want to devise a method of feature description which is robust with respect to the spatial layout of objects in a scene.
Although commonly used local pooling layers (max or mean pooling) in  standard CNN architectures have been shown to achieve viewpoint and pose invariance to some extent \cite{krizhevsky2012imagenet,he2014spatial}, these layers cannot accommodate large-scale deformations that are caused by spatial layout variations in indoor scenes.
In order to achieve spatial layout invariance, this paper introduces a modified CNN architecture with an additional layer, termed `spatially unstructured layer' (Sec. \ref{subsec:CNNArch}). The proposed CNN is then trained with images of indoor scenes (using our proposed strategy described in Sec.~\ref{Adapting Large-scale CNNs for Indoor Scene Categorization}) and the learnt feature representations are invariant to the spatial layout of the constituent objects.


Training a deep CNN requires a large amount of data because the number of parameters to be learnt is quite huge.
However, for the case of indoor scenes, we only have a limited number of annotated training data. This becomes then a serious limitation for the feasible training of a deep CNN.
Some recently proposed techniques demonstrate that pre-trained CNN models (on large datasets e.g., ImageNet) can be adapted for similar tasks with limited additional training data \cite{Chatfield14}.
However, cross domain adaptation becomes problematic in the case of heterogeneous tasks due to the different natures of source and target datasets.
For example, an image in the ImageNet dataset contains mostly centered objects belonging to only one class.
In contrast, an image in an indoor scene dataset has many constituent objects, all appearing in a variety of layouts and scales.
In this work, we propose an efficient strategy to achieve cross domain adaptation with only a limited number of annotated training images in the target dataset (Sec.~\ref{Adapting Large-scale CNNs for Indoor Scene Categorization}).


The major contributions of this paper can be summarized as: \textbf{1)} {\color{black}A new method of feature description (using the activations of a deep convolutional neural network) is proposed to deal with the large-scale spatial layout deformations in scene images } (Sec~\ref{subsec:CNNArch}),
\textbf{2)} A pyramidal image representation is proposed to achieve scale invariance (Sec~\ref{Pyramid Image Representation}), \textbf{3)} A novel transfer learning approach is introduced to efficiently adapt a pre-trained network model (on a large dataset) to any target classification task with only a small amount of available annotated training data (Sec~\ref{Adapting Large-scale CNNs for Indoor Scene Categorization}) and
\textbf{4)} Extensive experiments are performed to validate the proposed approach. Our results show a significant performance improvement for the challenging indoor scene classification task on a number of datasets.

\section{Related Work}
Indoor scene classification has been actively researched and a number of methods have been developed in recent years \cite{wu2011centrist,quattoni2009recognizing,pandey2011scene,lazebnik2006beyond,sun2013learning,razavian2014cnn,singh2012unsupervised,zuo2014learning}.
While some of these methods focus on the holistic properties of scene images (e.g., CENTRIST \cite{wu2011centrist}, Gist descriptor \cite{oliva2001modeling}), others give more importance to the local distinctive aspects (e.g., dense SIFT \cite{lazebnik2006beyond}, HOG \cite{xiao2010sun}).
In this paper, we argue that we cannot rely on either of the local or holistic image characteristics to describe all indoor scene types \cite{quattoni2009recognizing}.
For some scene types, holistic or global image characteristics are enough (e.g., \emph{corridor}), while for others, local image properties must be considered (e.g., \emph{bedroom}, \emph{shop}).
We therefore neither focus on the global nor the local feature description and instead extract mid-level image patches to encode an intermediate level of information. Further, we propose a pyramidal image representation which is able to capture the discriminative aspects of indoor scenes at multiple levels.


Recently, mid-level representations have emerged as a competitive candidate for indoor scene classification.
Strategies have been devised to discover discriminative mid-level image patches which are then encoded by a feature descriptor.
For example, the works \cite{juneja2013blocks, doersch2013mid, sun2013learning} learn to discover discriminative patches from the training data.
Our proposed method can also be categorized as a mid-level image patches based approach.
However, our method is different from previous methods, which require discriminative patch ranking and selection procedures or involve the learning of distinctive primitives.
In contrast, our method achieves state of the art performance by simply extracting mid-level patches densely and uniformly from an image (see more details in Sec.~\ref{Image Representation and Classification}.

An open problem in indoor scene classification is the design of feature descriptors which are robust to global layout deformations.
The initial efforts to resolve this problem used bag-of-visual-words models or variants (e.g., \cite{lazebnik2006beyond, bosch2008scene, yang2009linear}), which are based on locally invariant descriptors e.g., SIFT \cite{lowe2004distinctive}.
Recently, these local feature representations have been outperformed by learned feature representations from deep neural networks \cite{krizhevsky2012imagenet, ILSVRCarxiv14, razavian2014cnn}.
However, since there is no inherent mechanism in these deep networks to deal with the high variability of indoor scenes, several recent efforts have been made to fill in this gap (e.g., \cite{GongMOP14, he2014spatial}).
The bag of features approach of Gong et al. \cite{GongMOP14} performs VLAD pooling \cite{jegou2010aggregating} of CNN activations.
 Another example is the combination of spatial pyramid matching and CNNs (proposed by He et al. \cite{he2014spatial}) to increase the feature's robustness.
These methods, however, devise feature representations on top of CNN activations and do not inherently equip the deep architectures to effectively deal with the large deformations.
In contrast, this work provides an alternative strategy based on an improved network architecture to enhance invariance towards large scale deformations. The detailed description of our proposed feature representation method is presented next.

\begin{figure*}[t]
\centering
\includegraphics[width=1\textwidth]{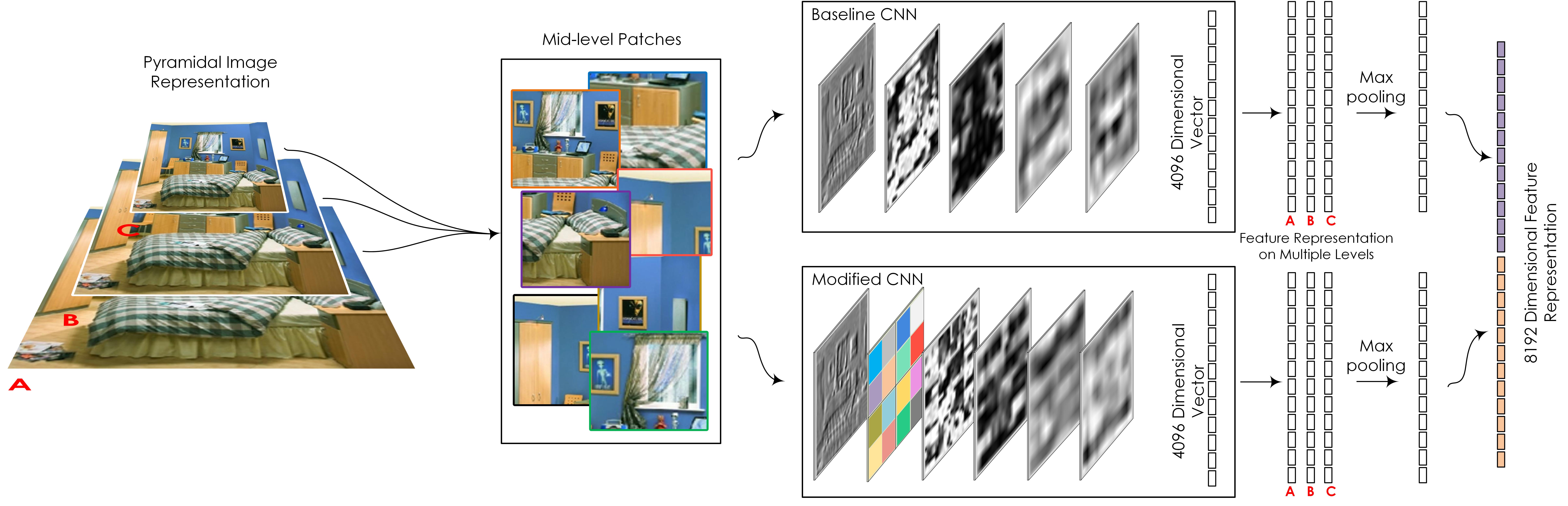}
\caption{Overview of the proposed Spatial Layout and Scale Invariant Convolutional Activations ($\text{S}^2\text{ICA}$) based feature description method. Mid-level patches are extracted from three levels (A, B, C) of the pyramidal image representation. The extracted patches are separately feed-forwarded to the two trained CNNs (with and without the spatially unstructured layer). The convolutional activations based feature representation of the patches is then pooled and a single feature vector for the image is finally generated by concatenating the feature vectors from both CNNs. Figure best seen in color.}
\label{fig:MainMethod}
\end{figure*}

\section{Proposed Spatial Layout and Scale Invariant Convolutional Activations - $\text{S}^2\text{ICA}$}
\label{Proposed Spatial Layout and Scale Invariant Convolutional Activations}
The block diagram of our proposed Spatial Layout and Scale Invariant Convolutional Activations ($\text{S}^2\text{ICA}$) based feature description method is presented in Fig~\ref{fig:MainMethod}. The detailed description of each of the blocks is given here. We first present our baseline CNN architecture followed by a detailed description of our spatially unstructured layer in Sec.~\ref{subsec:CNNArch}. Note that the spatially unstructured layer is introduced to achieve invariance to large scale spatial deformations, which are commonly encountered in images of indoor scenes. The baseline CNN architecture is pre-trained for a large scale classification task. A novel method is then proposed to adapt this pre-trained network for the specific task of scene categorization (Sec.~\ref{Adapting Large-scale CNNs for Indoor Scene Categorization}). Due to the data hungry nature of CNNs, it is not feasible to train a deep architecture with only a limited amount of available training data.
For this purpose, we pre-train a `TransferNet', which is then appended with the initialized CNN and the whole network can then be efficiently fine-tuned for the scene classification task. Convolutional activations from this fine-tuned network are then used for a robust feature representation of the input images. To deal with the scale variations, we propose a pyramidal image representation and combine the activations from multiple levels which results in a scale invariant feature representation (Sec.~\ref{Pyramid Image Representation}). This representation is then finally used by a linear Support Vector Machine (SVM) for classification (Sec.~\ref{Image Representation and Classification}).

\subsection{CNN Architecture}
\label{subsec:CNNArch}
\begin{figure*}[t]
\centering
\includegraphics[width=0.8\textwidth]{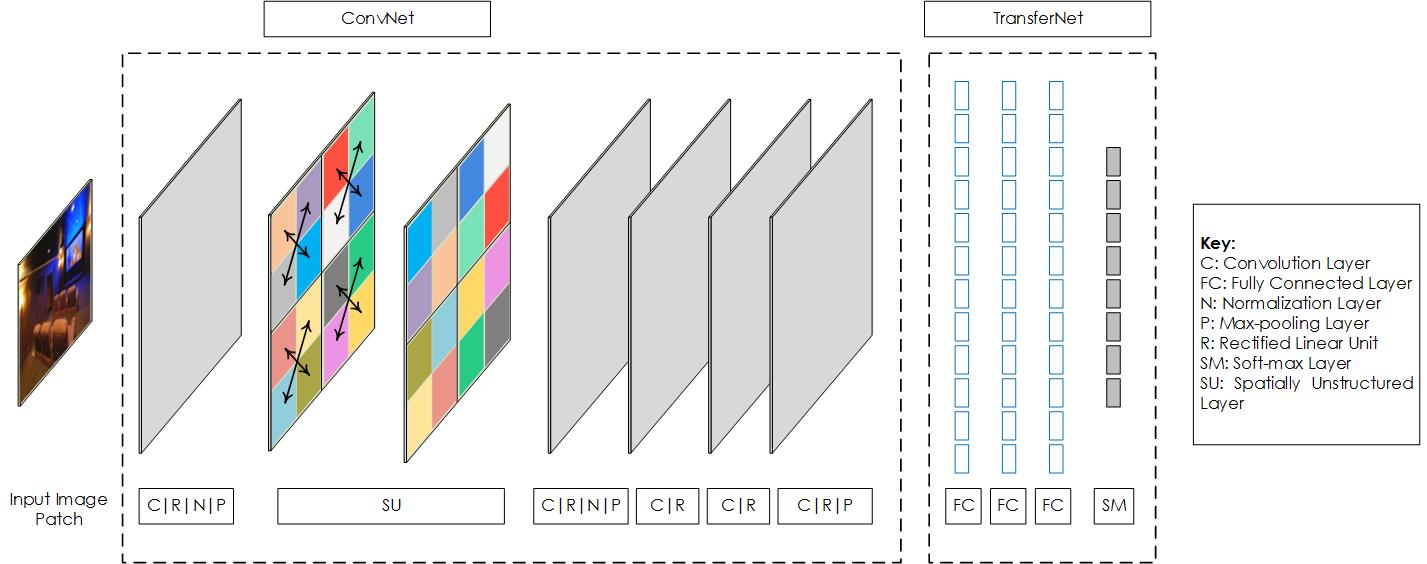}
\caption{The architecture of our proposed Convolutional Neural Network used to learn tailored feature representations for scene categorization. We devise a strategy (see Sec.~\ref{Adapting Large-scale CNNs for Indoor Scene Categorization} and Alg.~\ref{alg:s-svm}) to effectively adapt the learned feature representation from a large scale classification task to scene categorization.}
\label{fig:CNNArch}
\end{figure*}

Our baseline CNN architecture is presented in Fig~\ref{fig:CNNArch}. It  consists of five convolutional layers and four fully connected layers.
The architecture of our baseline CNN is similar to AlexNet \cite{krizhevsky2012imagenet}.
 The main difference is that we introduce extra fully connected layer, and that all of our neighboring layers are densely connected (in contrast to the sparse connections in AlexNet).
 To achieve spatial layout invariance, the architecture of the baseline CNN is modified and a new unstructured layer is added after the first sub-sampling layer.
A brief description of each layer of the network follows next.

Let us suppose that the convolutional neural network consists of $L$ hidden layers and each layer is indexed by $l \in \{ 1 \ldots L \}$.
The feed-forward pass can be described as a sequence of convolution, optional sub-sampling and normalization operations.
The response of each convolution node in layer $l$ is given by:
\begin{equation}
\mathbf{a}^{l}_{n} = f\left(\sum\limits_{m} (\mathbf{a}^{l-1}_{m} * \mathbf{k}^{l}_{m,n}) + b^{l}_{n}\right),
\end{equation}
where $\mathbf{k}$ and $b$ denote the learned kernel and  bias, the indices $(m,n)$ indicate that the mapping is from the $m^{th}$ feature map of the previous layer to the $n^{th}$ feature map of the current layer.
The function $f$ is the element-wise Rectified Linear Unit (ReLU) activation function.
The response of each normalization layer is given by:
\begin{equation}
\mathbf{a}_n^{l} = \frac{\mathbf{a}_n^{l-1}}{\left(\alpha + \beta \sum\limits_{j = max(0, n - \sigma)}^{min(N-1, n+\sigma)} (\mathbf{a}^{l-1}_{j} )^2 \right)^{\gamma}},
\end{equation}
where $\alpha, \beta, \gamma, \sigma$\footnote{These constants are defined as in \cite{krizhevsky2012imagenet}: $\alpha = 2$, $\beta = 10^{-4}$, $\gamma=3/4$ and $\sigma = 5/2$.} are constants and $N$ is the total number of kernels in the layer.
The response of each sub-sampling node is given by:
\begin{equation}
\mathbf{a}^{l}_{n} = \frac{k^l_n}{T^2}\sum\limits_{T \times T}\mathbf{a}^{l-1}_n + b^l_n
\end{equation}
where, $k_n^l$ is the connection weight and $T$ is the neighborhood size over which the values are pooled.

\begin{figure}
\centering
\includegraphics[width=\columnwidth]{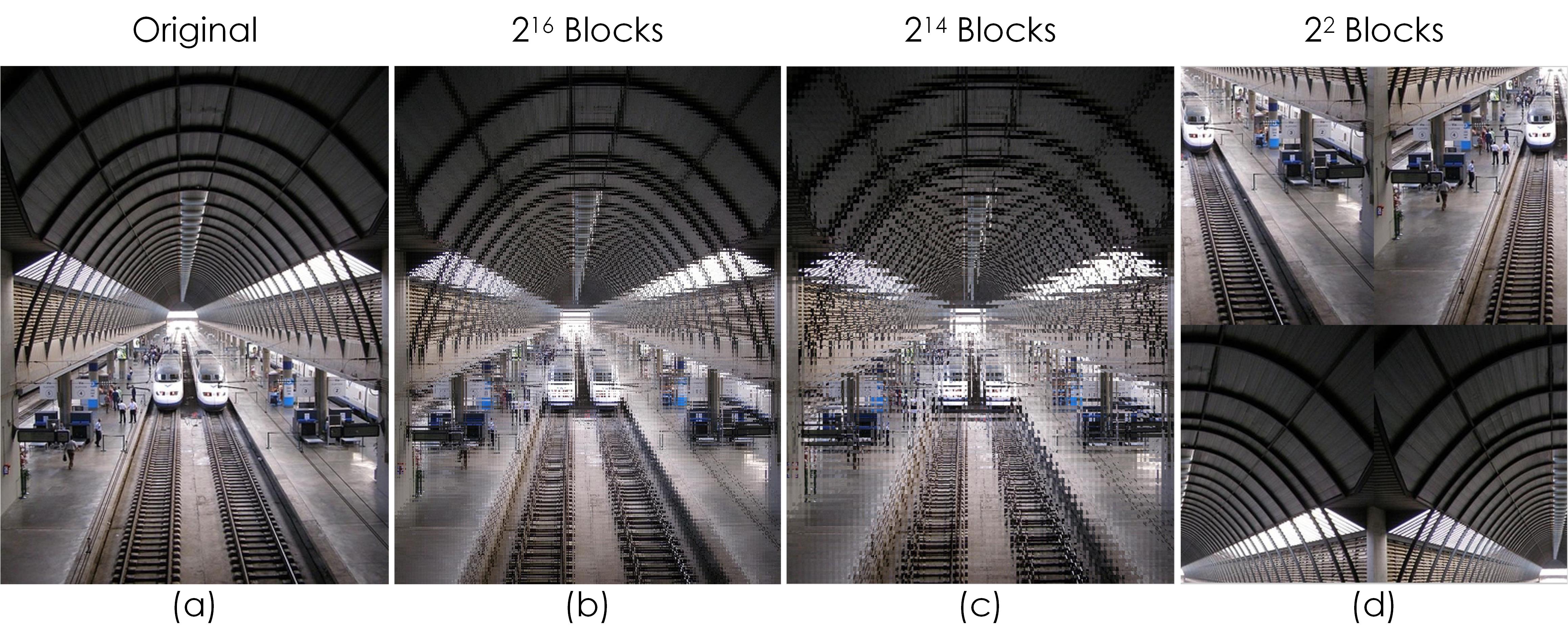}
\caption{(\emph{left} to \emph{right}) Original image and the spatially unstructured versions with $2^{16}$, $2^{14}$ and $2^{2}$ blocks respectively. }
\label{fig:distortion}
\end{figure}

In our proposed modified CNN architecture, a spatially unstructured layer follows the first sub-sampling layer and breaks the spatial order of the output feature maps. This helps in the generation of robust feature representations that can cope with the high variability of indoor scenes.
For each feature response, we split the feature map into a specified number of blocks ($n$).
Next, a matrix $\mathbf{U}$ is constructed whose elements correspond to the scope of each block defined as a tuple:
\begin{equation}
\mathbf{U}_{\sqrt{n}\times\sqrt{n}} = \{\mathbf{u}_i \; \forall i \;| \mathbf{u}_i =  (p,q) \},
\end{equation}
where, $p$ and $q$ indicate the starting and ending index of each block.
To perform a local swapping operation, we define a matrix $\mathbf{S}$ in terms of an identity matrix $I$ as follows:
\begin{equation}
\mathbf{S}_{2\times2} = |I - 1| =
\left(
\begin{array}{cc}
0 & 1 \\
1 & 0
\end{array} \right)
\end{equation}
Next, a transformation matrix $\mathbf{T} \in \mathbb{R}^{\sqrt{n}\times\sqrt{n}}$ is defined in terms of $\mathbf{S}$ as follows:
\begin{equation}
\mathbf{T}_{\sqrt{n}\times\sqrt{n}} =
\left(
\begin{array}{cccc}
\mathbf{S} &  \mathbf{0} & \ldots & \mathbf{0} \\
\mathbf{0} &  \mathbf{S} &  \ldots & \mathbf{0} \\
\vdots     &  \vdots     & \ddots & \vdots     \\
\mathbf{0} &  \mathbf{0} & \ldots & \mathbf{S}
\end{array}
\right)_{\sqrt{n}/2\times\sqrt{n}/2}
\end{equation}
The transformation matrix $\mathbf{T}$ has the following properties:
\begin{itemize}
\itemsep0em
\item  $\mathbf{T} = \{t_{ij}\}$ is a permutation matrix ($\mathbf{T}: \{\mathbf{u}_{ij}\} \rightarrow \{\mathbf{u}_{ij}\} $) since the sum along each row and column is always equal to one i.e., $\sum\limits_{i} t_{ij} = \sum\limits_{j} t_{ij} = 1$.

\item $\mathbf{T}$ is a bistochastic matrix and therefore according to Birkhoff–von Neumann theorem and the above property,  $\mathbf{T}$ lies on the convex hull of the set of bistochastic matrices.
\item  It is a binary matrix with entries belonging to the Boolean domain $\{0,1\}$.

\item Its an orthogonal matrix i.e., $\mathbf{T} \mathbf{T}^{T} = I$ and $\mathbf{T}^{-1} = \mathbf{T}^{T}$.
\end{itemize}
Using the matrix $\mathbf{T}$, we transform $\mathbf{U}$ to become:
\begin{equation}
\hat{\mathbf{U}} = (\mathbf{U}^{T}\mathbf{T})^{T}\mathbf{T} = \mathbf{T}^{T} \mathbf{U} \mathbf{T}.
\end{equation}
The updated matrix $\hat{\mathbf{U}} $ contains the new indices of the modified feature maps.
If $\mathcal{Y}(\cdot)$ is a function which reads the indices of the blocks stored in the form of tuples in matrix $\hat{\mathbf{U}}$, the layer output are as follows:
\begin{equation}
\mathbf{a}_{n}^{l} = r * \mathcal{Y}(\mathbf{a}_{n}^{l-1}, \hat{\mathbf{U}}),
\end{equation}
\begin{equation}
\text{where, } r \sim \text{Bernoulli}(p).
\end{equation}
$r$ is a random variable which has a probability $p$ of being equal to $1$.
Note that this shuffling operation is applied randomly so that a network does not get biased towards the normal patches.
Fig.~\ref{fig:distortion} illustrates the distortion operation performed by the spatially unstructured layer for a different number of blocks.

\begin{algorithm*}[htp]
  \caption{Operations Involved in Spatially Unstructured Layer}
  \begin{algorithmic}
  \Require
  Feature map : $\mathbf{F} \in \mathbf{M}_{p\times q \times r \times s}(\mathbb{R})$, Number of Blocks : $n$

  // $\mathbf{F}$ is a real valued four dimensional matrix

  \Ensure Modified feature map ($ \mathbf{F}_m$)

  \State $\ell = \floor*{\frac{\sqrt{n}}{2}}$ \hfill // Rearrangement level
  \State $h_{pts} \leftarrow (\ell + 1)$ linearly spaced points in range $[1:p]$
  \State $h_{pts} = \floor*{h_{pts}}$
  \State $h_{pts}[end] \; += 1$
  \State $w_{pts} = h_{pts}$  \hfill // $\because p = q \text{ for } \mathbf{F}$
  \For {$ \forall i \in [1:length(h_{pts})-1]$}
  	\For {$ \forall j \in [1:length(w_{pts})-1]$}
  	
  	\State \resizebox{0.6\textwidth}{!}{$F_{tmp} = \mathbf{F}[h_{pts}(i):h_{pts}(i+1)-1, w_{pts}(j):   w_{pts}(j+1) -1,:,:] $}
  	
  	\State \resizebox{0.65\textwidth}{!}{$F_{tmp} =  [F_{tmp}(\ceil*{\frac{rows(F_{tmp})}{2}}: end, :, :, :) ;  F_{tmp}(1: \floor*{\frac{rows(F_{tmp})}{2}}, :, :, :)]$}
  	
  	\State \resizebox{0.63\textwidth}{!}{$F_{tmp} =  [F_{tmp}(:, \ceil*{\frac{cols(F_{tmp})}{2}}: end,  :, :) ;  F_{tmp}(:, 1:   \floor*{\frac{cols(F_{tmp})}{2}},  :, :)]   $}  	
  	
  	\State \resizebox{0.65\textwidth}{!}{$\mathbf{F}_m [h_{pts}  (i):h_{pts}(i+1)-1, w_{pts}(j):w_{pts}(j+1) -1,  :,:] = F_{tmp}$}

	\EndFor
  \EndFor

  \State \Return $\{\mathbf{R}\}$
  \end{algorithmic}\label{alg:region_growing}
\end{algorithm*}

\subsection{Training CNNs for Indoor Scenes}
\label{Adapting Large-scale CNNs for Indoor Scene Categorization}

Deep CNNs have demonstrated exceptional feature representation capabilities for the classification and detection tasks (e.g., see ILSVRC'14 Results \cite{ILSVRCarxiv14}). Training deep CNNs however requires a large amount of data since the number of parameters to be learnt is huge. The requirement of a large amount of training  data makes the training of CNNs infeasible where only a limited amount of annotated training data is available. In this paper, we propose to leverage from the image representations learnt on a large scale classification task (such as on ImageNet \cite{ILSVRCarxiv14}) and propose a strategy to learn tailored feature representations for indoor scene categorization. An algorithmic description of our proposed strategy is summarized in Algorithm.~\ref{alg:s-svm}. The details are presented here.

We first train our baseline CNN architecture on ImageNet database following the procedure in \cite{krizhevsky2012imagenet}. Next, we densely extract mid-level image patches from our scene classification training data and represent them in terms of the convolutional activations of the trained baseline network. The output of the last convolution layer followed by ReLU non-linearity is considered as a feature representation of the extracted patches. These feature representations ($\mathcal{F}$) will be used to train our TransferNet.

As depicted in Fig~\ref{fig:CNNArch}, our TransferNet consists of three hidden layers (with 4096 neurons each) and an output layer, whose number of neurons are equal to the number of classes in the target dataset (e.g., indoor scenes dataset). TransferNet is trained on convolutional feature representations ($\mathcal{F}$) of mid-level patches of the scene classification dataset.
Specifically, the input to TransferNet are the feature representations ($\mathcal{F}$) of the patches and the outputs are their corresponding class labels.
After training TransferNet, we remove all fully connected layers of the baseline CNN and join the trained TransferNet to the last convolutional layer of the baseline CNN.
The resulting network then consists of five convolutional layers and four fully connected layers (of the trained TransferNet).
This complete network is now fine-tuned on the patches extracted from the training images of the scene classification data.
Since the network initialization is quite good (the convolutional layers of the network are initialized from the baseline network trained on imageNet dataset, whereas the fully connected layers are initialized from the trained transferNet), only few epochs are required for the network to converge.
Moreover, with a good initialization, it becomes feasible to learn deep CNN's parameters even with a smaller number of available training images.

Note that the baseline CNN was trained with images from the ImageNet database, where each image pre-dominantly contains one or multiple instances of the same object.
In the case of scene categorization, we may deal with multiple distinct objects from a wide range of poses, appearances and scales across different spatial locations.
Therefore, in order to incorporate large scale deformations, we train two CNNs: with and without the spatially unstructured layer (learned weights represented by $\mathbf{W}$ and $\mathbf{W}_{su}$ respectively).
These trained CNNs are then used for robust feature representation in Sec.~\ref{Image Representation and Classification}. Below, we first explain our approach in dealing with scale variations.


 \begin{algorithm}[h]
  \caption{Training CNNs for indoor scenes}
  \begin{algorithmic}[1]
  \Require
  Source DB (ImageNet), Target DB (Scene Images)
  \Ensure
  Learned weights: $\{\mathbf{W}\}_{1\times L}$, $\{\mathbf{W}_{su}\}_{1\times L}$
  \State Pre-train the CNN on the large-scale Source DB.
  \State Feed-forward image patches from target DB to trained CNN.
  \State Take feature representations ($\mathcal{F}$) from the last convolution layer.
  \State Train the `TransferNet' of $4$ fully connected layers with $\mathcal{F}$ as input and target annotations as output.
  \State Append `TransferNet' to the last convolution layer of trained CNN.
  \State Fine-tune the complete network with and without the spatially unstructured layer to get $\{\mathbf{W}\}_{1\times L}$ and $\{\mathbf{W}_{su}\}_{1\times L}$ respectively.

  \end{algorithmic}\label{alg:s-svm}
\end{algorithm}

\subsection{Pyramid Image Representation}
\label{Pyramid Image Representation}
In order to achieve scale invariance, we generate a pyramid of an image at multiple spatial resolutions.
However, unlike conventional pyramid generation processes (e.g., Gaussian or Laplacian pyramid) where smoothing and sub-sampling operations are repeatedly applied, we simply resize each image to a set of scales and this may involve up or down sampling.
Specifically, we transform each image to three scales, $\{0.75 \times D, D, 1.25 \times D \}$, where $D$ is the smaller dimension of an image which is set based on the given dataset.
At each scale, we densely extract patches which are then encoded in terms of the convolutional activations of the trained CNNs.

\subsection{Image Representation and Classification}
\label{Image Representation and Classification}
From each of the three images of the pyramidal image representation, we extract multiple overlapping patches of $224 \times 224$ using a sliding window. A shift of $32$ pixels is used between patches. The extracted image patches are then fed forwarded to the trained CNNs (both with and without the spatially unstructured layer). The convolutional feature representation of the patches are max-pooled to get a single feature vector representation for the image. This is denoted by A, B and C corresponding to three images of the pyramid in Fig~\ref{fig:MainMethod}. We then max pool the feature representations of these images and generate one single representation of the image for each network (with and without the spatially unstructured layer). The final feature representation is achieved by concatenating these two feature vectors. After encoding the spatial layout and the scale invariant feature representations for the images, the next step is to perform classification. We use a simple linear Support Vector Machine (SVM) classifier for this purpose.

\begin{figure*}[pth]
\centering
\includegraphics[width=.8\linewidth]{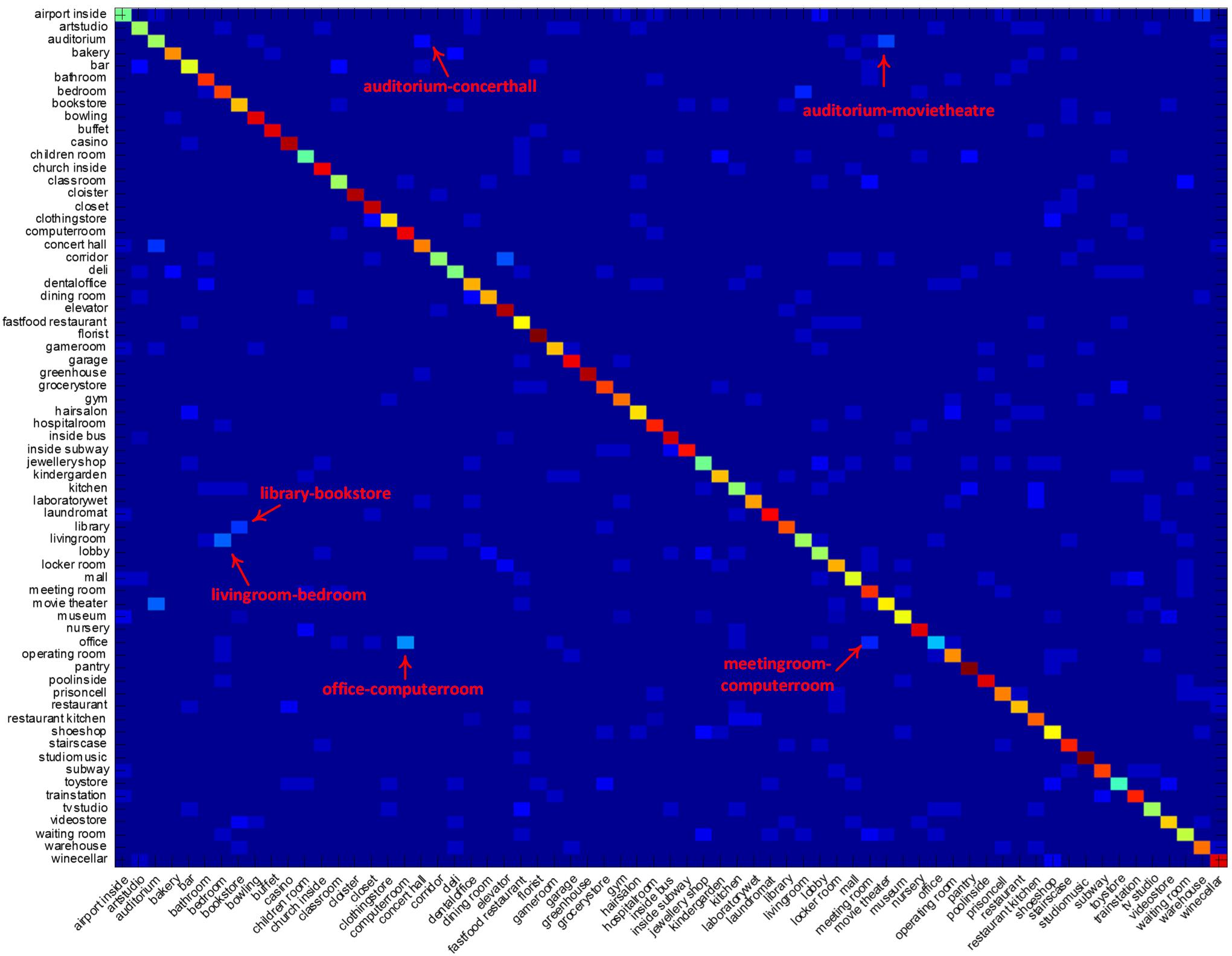}
\caption{Confusion Matrix for the MIT 67 Indoor Scenes Dataset. Figure best seen in color.}
\label{fig:confu_mat_MIT}
\end{figure*}

\begin{figure*}[tph]
\centering
\begin{subfigure}[]{0.27\textwidth}
\imagebox{45mm}{\includegraphics[width=\textwidth]{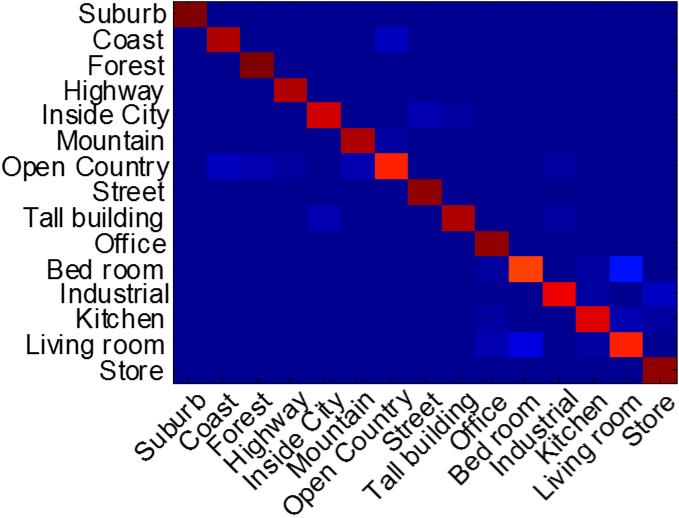}}
\label{1}
\end{subfigure}
\;
\begin{subfigure}[]{0.30\textwidth}
\imagebox{50mm}{\includegraphics[trim= 0 0 4em 0, clip, width=\textwidth]{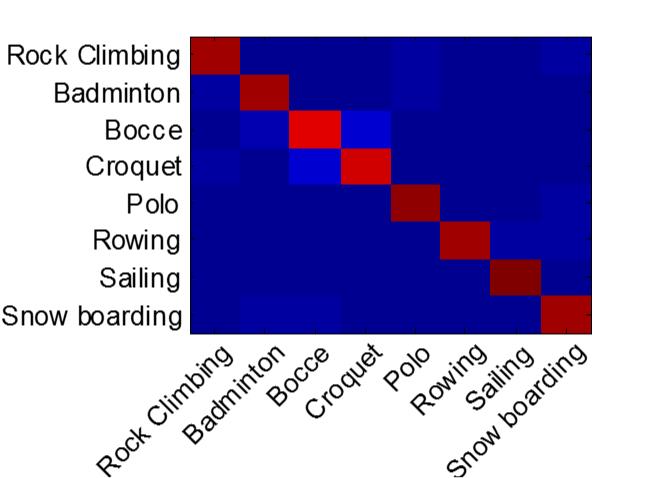}}
\label{2}
\end{subfigure}
\;
\begin{subfigure}[]{0.34\textwidth}
\imagebox{50mm}{\includegraphics[trim= 0 0 0em 0, clip, width=\textwidth]{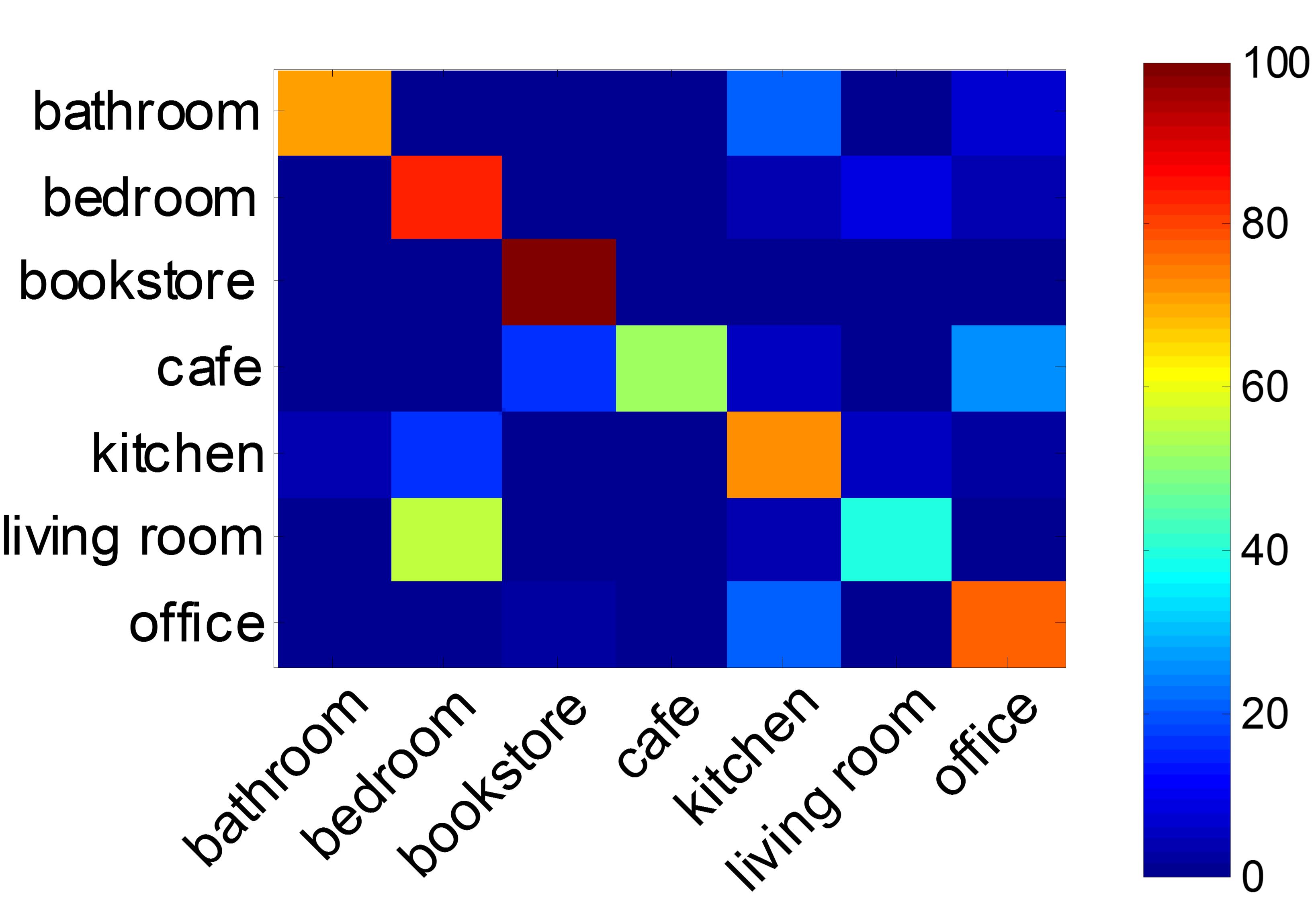}}
\label{3}
\end{subfigure}
\caption{Confusion matrices for Scene-15, Sports-8 and NYU scene classification datasets. Figure best seen in color.}
\label{fig:confu_mat}
\end{figure*}

\section{Experiments and Evaluation}
The proposed approach is validated through extensive experiments on a number of datasets. To this end, we perform experiments on three indoor scene datasets (MIT-67, NYU and Scene-15). Amongst these datasets, MIT-67 is the largest dataset for indoor scene classification. The dataset is quite challenging since images of many classes are similar in appearance and thus hard to classify (see Fig.~\ref{fig: misclassImages}). Apart from indoor scene classification, we further validate our approach for two other tasks i.e., event and object datasets (Graz-02 and Sports-8). Below (Sec.~\ref{Datasets}), we first present a brief description about each of the datasets and the adopted experimental protocols. We then present our experimental results along with a comparison with existing state of the art in Sec.~\ref{subsec:results}. An ablative analysis to study the individual effect of each component on the proposed method is also presented in Sec.~\ref{subsec:results}.

\begin{table*}[pth]
\centering
\begin{tabular}{l|c|| l|c}
\toprule
 \multicolumn{4}{c}{\textbf{MIT-67 Indoor Scenes Dataset}} \\
 \midrule
Method & Accuracy(\%) & Method & Accuracy (\%) \\
\hline\hline
ROI + GIST [CVPR'09] \cite{quattoni2009recognizing} & $26.1$ & OTC [ECCV'14] \cite{margolin2014otc} & $47.3$ \\

MM-Scene [NIPS'10] \cite{zhu2010large} & $28.3$  & Discriminative Patches [ECCV'12] \cite{singh2012unsupervised}
& $49.4$ \\

SPM [CVPR'06] \cite{lazebnik2006beyond} & $34.4$ & ISPR  [CVPR'14]\cite{linlearning14} & $50.1$ \\

Object Bank [NIPS'10] \cite{li2010object} & $37.6$& D-Parts [ICCV'13] \cite{sun2013learning} & $51.4$ \\

RBoW [CVPR'12] \cite{parizi2012reconfigurable} & $37.9$ & VC + VQ [CVPR'13] \cite{li2013harvesting} & $52.3$ \\

Weakly Supervised DPM [ICCV'11] \cite{pandey2011scene} & $43.1$  & IFV [CVPR'13]\cite{juneja2013blocks}  & $60.8$ \\

SPMSM [ECCV'12] \cite{kwitt2012scene} & $44.0$ & MLRep [NIPS'13] \cite{doersch2013mid} & $64.0$ \\

LPR-LIN [ECCV'12] \cite{sadeghi2012latent} & $44.8$ & CNN-MOP [ECCV'14]\cite{GongMOP14}  & $68.9$ \\

BoP [CVPR'13] \cite{juneja2013blocks} & $46.1$ & CNNaug-SVM [CVPRw'14] \cite{razavian2014cnn} & $69.0$ \\
\cline{3-4}
Hybrid Parts + GIST + SP [ECCV'12] \cite{zheng2012learning} & $47.2$ & \textbf{Proposed $\text{S}^2\text{ICA}$} & \textbf{71.2}\\

\bottomrule
\end{tabular}
\caption{Mean accuracy on the MIT-67 indoor scenes dataset.}
\label{tab:MIT67acc}
\end{table*}



\subsection{Datasets}
\label{Datasets}

\noindent
The \textbf{MIT-67 Dataset}
 contains a total of 15620 images of 67 indoor scene classes. For our experiments, we follow the standard evaluation protocol in \cite{quattoni2009recognizing}. Specifically, 100 images per class are considered, out of which 80 are used for training and the remaining 20 are used for testing. We therefore have a total of 5360 and 1340 images for training and testing respectively. 

\noindent
The \textbf{15 Category Scene Dataset} contains images of 15 urban and natural scene classes. The number of images for each scene class in the dataset ranges from 200-400. For performance evaluation and comparison with existing state of the art, we follow the standard evaluation protocol in \cite{lazebnik2006beyond}, where 100 images per class are selected for training and the rest for testing.

\noindent
The \textbf{NYU v1 Indoor Scene Dataset} contains a total of 2347 images belonging to 7 indoor scene categories. We follow the evaluation protocol described in \cite{silberman11indoor} and use the first $60\%$ of the images of each class for training and the last $40\%$ images for testing.

\noindent
The \textbf{Inria Graz 02 Dataset} contains a total of 1096 images of three classes (bikes, cars and people). The images of this dataset exhibit a wide range of appearance variations in the form of heavy clutter, occlusions and pose changes. The evaluation protocol defined in \cite{marszatek2007accurate} is used in our experiments. Specifically, the training and testing splits are generated by considering the first 150 odd images for training and the first 150 even images for testing.

\noindent
The \textbf{UIUC Sports Event Dataset}
contains 1574 images of 8 sports event categories. Following the protocol defined in \cite{li2007and}, we used $70$ and $60$ randomly sampled images per category for training and testing respectively.


\subsection{Results and Analysis}\label{subsec:results}

The quantitative results of the proposed method in terms of classification rates for the task of indoor scene categorization are presented in Tables~\ref{tab:MIT67acc},~\ref{tab:nyu_acc}~and~\ref{tab:15scenes_acc}. A comparison with the existing state of the art techniques shows that the proposed method consistently achieves a superior performance on all datasets. We also evaluate the proposed method for the tasks of sports events and highly occluded object classification (Tables~\ref{tab:Sports_acc}~and~\ref{tab:graz_acc}). The results show that the proposed method achieves very high classification rates. The experimental results suggest that the gain in performance of our method is more significant and pronounced for the MIT-67, Scene-15, Graz-02 and Sports-8 datasets.
The confusion matrices showing the class wise accuracies of Scene-15, Sports-8 and NYU datasets are presented in Fig.~\ref{fig:confu_mat}.
The confusion matrix for the MIT-67 scene dataset is given in Fig.~\ref{fig:confu_mat_MIT}.
It can be noted that all the confusion matrices have a very strong diagonal (Fig.~\ref{fig:confu_mat_MIT} and Fig.~\ref{fig:confu_mat}). The majority of the confused testing samples belong to very closely related classes e.g., \emph{living room} is confused with \emph{bedroom}, \emph{office} with \emph{computer-room}, \emph{coast} with \emph{open-country} and \emph{croquet} with \emph{bocce}.

\begin{figure}
\centering
\includegraphics[width=\columnwidth]{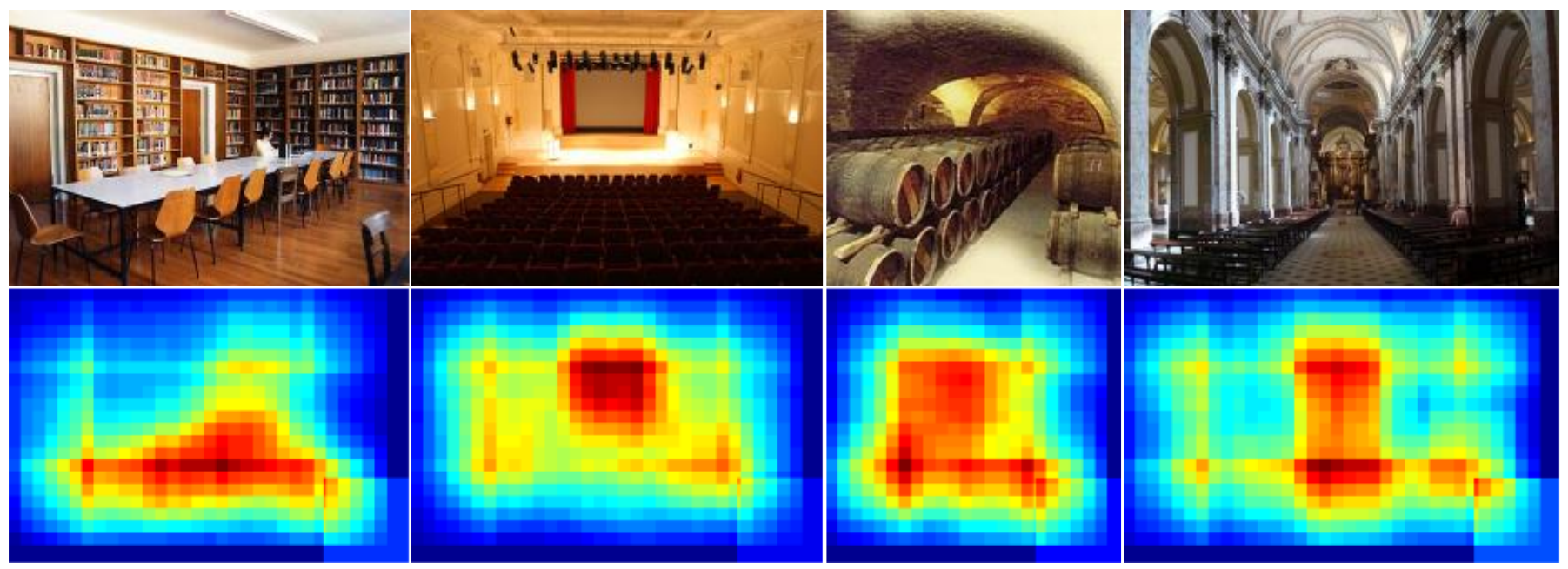}
\caption{The contributions (\emph{red}: most; \emph{blue}: least) of mid-level patches towards correct class prediction. Best seen in color.}
\label{fig:DiscriminativePacthes}
\end{figure}

The superior performance of our method is attributed to its ability to handle a large spatial layout (through the introduction of the spatially unstructured layer in our modified CNN architecture) and scale variations (achieved by the proposed pyramidal image representation). Further, our method is based on deep convolutional representations, which have recently shown  to be superior in performance over shallow handcrafted feature representations \cite{razavian2014cnn,he2014spatial, ILSVRCarxiv14}.
A number of compared methods are based upon mid-level feature representations (e.g., \cite{juneja2013blocks, doersch2013mid, sun2013learning}). Our results show that our proposed method achieves superior performance over these methods. It should be noted that in contrast to existing mid-level feature representation based methods (whose main focus is on the automatic discovery of discriminative mid-level patches) our method simply densely extracts mid-level patches from uniform locations across an image. This is computationally very efficient since we do not need to devise patch selection and sorting strategies. Further, our dense patch extraction is similar to dense keypoint extraction, which has shown a comparable performance with sophisticated keypoint extraction methods over a number of classification tasks \cite{hayatHSI}. The contributions of the extracted mid-level patches towards a correct classification are shown in the form of heat maps for some example images in Fig~\ref{fig:DiscriminativePacthes}. It can be seen that our proposed spatial layout and scale invariant convolutional activations based feature descriptor gives automatically more importance to the meaningful and information rich parts of an image.

\begin{figure*}
\centering
\includegraphics[trim= 0 0 0 0, clip, width=\textwidth]{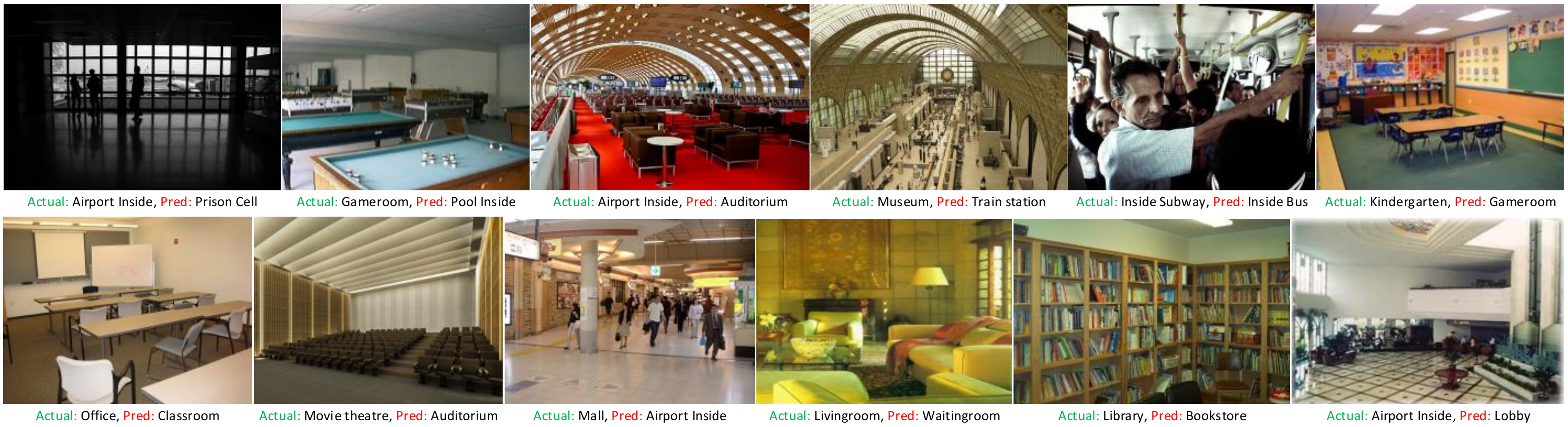}
\caption{Some examples of misclassified images from MIT-67 indoor scenes dataset. Actual and predicted labels of each image are given. Images from highly similar looking classes are confused amongst each other. For example, the proposed method misclassifies \emph{library} as \emph{bookstore}, \emph{office} as \emph{classroom} and \emph{inside subway} as \emph{inside bus}.}
\label{fig: misclassImages}
\end{figure*}

The actual and predicted labels of some miss-classified images from MIT-67 dataset are shown in Fig~\ref{fig: misclassImages}. Note the extremely challenging nature of the images in the presence of high inter-class similarities. Some of the classes are very challenging and there is no visual indication to determine the actual label. It can be seen that the miss-classified images belong to highly confusing and very similar looking scene types. For example, the image of \emph{inside subway} is miss-classified as \emph{inside bus}, \emph{library} as \emph{bookstore}, \emph{movie theater} as \emph{auditorium} and \emph{office} as \emph{classroom}.

\begin{table}
\centering
\begin{tabular}{l|c}
\toprule
 \multicolumn{2}{c}{\textbf{UIUC Sports-8 Dataset}} \\
 \midrule
Method & Accuracy (\%) \\
\hline\hline
GIST-color [IJCV'01] \cite{oliva2001modeling} & $70.7$ \\

MM-Scene [NIPS'10] \cite{zhu2010large} & $71.7$ \\

Graphical Model [ICCV'07] \cite{li2007and} & $73.4$ \\

Object Bank [NIPS'10] \cite{li2010object} & $76.3$ \\

Object Attributes [ECCV'12] \cite{li2012objects} & $77.9$ \\

CENTRIST [PAMI'11] \cite{wu2011centrist} &  $78.2$ \\

RSP [ECCV'12] \cite{jiang2012randomized} & $79.6$ \\

SPM [CVPR'06] \cite{lazebnik2006beyond} & $81.8$ \\

SPMSM [ECCV'12] \cite{kwitt2012scene} & $83.0$ \\

Classemes [ECCV'10] \cite{torresani2010efficient} & $84.2$ \\

HIK [ICCV'09] \cite{wu2009beyond} & $84.2$ \\

LScSPM [CVPR'10] \cite{gao2010local} & $85.3$ \\

LPR-RBF [ECCV'12] \cite{sadeghi2012latent} & $86.2$ \\

Hybrid Parts + GIST + SP [ECCV'12] \cite{zheng2012learning} & $87.2$ \\

LCSR [CVPR'12] \cite{shabou2012locality} &  $87.2$ \\

VC + VQ [CVPR'13] \cite{li2013harvesting} & $88.4$ \\

IFV \cite{vedaldi08vlfeat} & $90.8$ \\

ISPR [CVPR'14] \cite{linlearning14} & $89.5$ \\
\hline
\textbf{Proposed $\text{S}^2\text{ICA}$} & \textbf{95.8} \\
\bottomrule
\end{tabular}
\caption{Mean accuracy on the UIUC Sports-8 dataset. }
\label{tab:Sports_acc}
\end{table}


\begin{table}
\centering
\begin{tabular}{l|c}
\toprule
 \multicolumn{2}{c}{\textbf{NYU Indoor Scenes Dataset}} \\
 \midrule
Method & Accuracy (\%) \\
\hline\hline
BoW-SIFT [ICCVw'11] \cite{silberman11indoor} & $55.2$ \\
RGB-LLC [TC'13] \cite{tao2013rank} & $78.1$ \\
RGB-LLC-RPSL [TC'13] \cite{tao2013rank} & $79.5$ \\
\hline
\textbf{Proposed $\text{S}^2\text{ICA}$} & $\mathbf{81.2}$\\
\bottomrule
\end{tabular}
\caption{Mean Accuracy for the NYU v1 dataset. }
\label{tab:nyu_acc}
\end{table}

\begin{table}
\centering
\begin{tabular}{l|ccc|c}
\toprule
 \multicolumn{5}{c}{\textbf{Graz-02 Dataset}} \\
 \midrule
& Cars & People & Bikes & Overall \\
\hline\hline
OLB [SCIA'05] \cite{opelt2005object} & 70.7 & 81.0
 & 76.5  & 76.1 \\
 VQ [ICCV'07] \cite{tuytelaars2007vector} & 80.2 & 85.2 & 89.5 & 85.0 \\
ERC-F [PAMI'08] \cite{moosmann2008randomized} & 79.9 & - & 84.4 & {82.1}\\
TSD-IB [BMVC'11] \cite{krapac2011learning} & 87.5 & 85.3 & 91.2 &  88.0 \\
TSD-k [BMVC'11] \cite{krapac2011learning} & 84.8 & 87.3 & 90.7 & 87.6 \\
\hline
\textbf{Proposed $\text{S}^2\text{ICA}$} & 98.7 & 97.7 &   97.7 & \textbf{98.0} \\
\bottomrule
\end{tabular}
\caption{Equal Error Rates (EER) on Graz-02 dataset. }
\label{tab:graz_acc}
\end{table}

\begin{table*}[th]
\centering
\begin{tabular}{l|c || l|c}
\toprule
 \multicolumn{4}{c}{\textbf{15 Category Scene Dataset}} \\
 \midrule
Method & Accuracy(\%) & Method & Accuracy (\%) \\
\hline\hline
 GIST-color [IJCV'01] \cite{oliva2001modeling} & $69.5$ &  ISPR [CVPR'14] \cite{linlearning14} & $85.1$ \\

RBoW [CVPR'12] \cite{parizi2012reconfigurable} & $78.6$ & VC + VQ [CVPR'13] \cite{li2013harvesting} & $85.4$ \\

Classemes [ECCV'10] \cite{torresani2010efficient} & $80.6$ & LMLF [CVPR'10] \cite{boureau2010learning} & $85.6$ \\

Object Bank [NIPS'10] \cite{li2010object} & $80.9$
& LPR-RBF [ECCV'12] \cite{sadeghi2012latent} & $85.8$ \\

SPM [CVPR'06] \cite{lazebnik2006beyond} & $81.4$ & Hybrid Parts + GIST + SP [ECCV'12] \cite{zheng2012learning} & $86.3$ \\

SPMSM [ECCV'12] \cite{kwitt2012scene} & $82.3$ &  CENTRIST+LCC+Boosting [CVPR'11] \cite{yuan2011mining}  & $87.8$ \\

LCSR [CVPR'12] \cite{shabou2012locality} &  $82.7$ & RSP [ECCV'12] \cite{jiang2012randomized} & $88.1$ \\

SP-pLSA [PAMI'08] \cite{bosch2008scene} & $83.7$ & IFV \cite{vedaldi08vlfeat} & $89.2$ \\

CENTRIST [PAMI'11] \cite{wu2011centrist} &  $83.9$ & LScSPM [CVPR'10] \cite{gao2010local} & $89.7$ \\

 HIK [ICCV'09]\cite{wu2009beyond} & $84.1$
& \\

\cline{3-4}
OTC [ECCV'14] \cite{margolin2014otc} & $84.4$
& \textbf{Proposed $\text{S}^2\text{ICA}$} & \textbf{93.1} \\
\bottomrule
\end{tabular}
\caption{Mean accuracy on the 15 Category scene dataset. Comparisons with the previous best techniques are also shown.}
\label{tab:15scenes_acc}
\end{table*}

An ablative analysis to assess the effect of each individual component of the proposed technique towards the overall performance is presented in Table~\ref{tab:ablation}.
Specifically, the contributions of the proposed spatially unstructured layer, pyramid image representation, training of the CNN on the target dataset and pooling (mean pooling and max pooling) are investigated. In order to investigate a specific componenet of the proposed method, we only modify (add or remove) that part, while the rest of the pipeline is kept fixed.
The experimental results in Table~\ref{tab:ablation} show that the feature representations from trained CNNs with and without the spatially unstructured layer complement each other and achieve the best performance. Furthermore, the proposed pyramidal image representation also contributes significantly towards the performance improvement of the proposed method. Our proposed strategy to adapt a deep CNN (trained on a large scale classification task) for scene categorization also proves to be very effective and it results in a significant performance improvement. Amongst the pooling strategies, max pooling provides a superior performance compared with mean pooling.


\begin{table}[tph]
\centering
\begin{tabular}{l|c}
Baseline CNN (w/o Spatially Unstructured layer)  & $65.4\%$ \\
Modified CNN (with Spatially Unstructured layer) & $65.9\%$ \\
Baseline CNN + Modified CNN  & $71.2\%$\\
\hline
w/o pyramidal representation & $68.5\%$\\
with pyramidal representation & $71.2\%$\\
\hline
\hline
CNN trained on imageNet & $67.3\%$\\
CNN trained on imageNet+MIT-67 & $71.2\%$\\
\hline
Mean-pooling & $65.7\%$\\
Max-pooling & $71.2\%$\\
\end{tabular}
\caption{Ablative analysis on MIT-67 dataset. The joint feature representations from baseline and modified CNNs gives the best performance. The proposed pyramidal image representation results in a significant performance boost.}
\label{tab:ablation}
\end{table}



\section{Conclusion}
This paper proposed a novel approach to handle the large scale deformations caused by spatial layout and scale variations in indoor scenes.
A pyramidal image representation has been contrived to deal with scale variations. A modified Convolutional Neural Network Architecture with an added layer has been introduced to deal with the variations caused by spatial layout changes. In order to feasibly train a CNN on tasks with only a limited annotated training dataset, the paper proposed an efficient strategy which conveniently transfers learning from a large scale dataset. A robust feature representation of an image is then achieved by extracting mid-level patches and encoding them in terms of the convolutional activations of the trained networks. Leveraging on the proposed spatial layout and scale invariant image representation, state of the art classification performance has been achieved by using a simple linear SVM classifier.

\section*{Acknowledgements}
This research was supported by the SIRF and IPRS scholarships from the University of Western Australia (UWA) and the Australian Research Council (ARC) grants DP110102166, DP150100294 and DP120102960.
We gratefully acknowledge the support of NVIDIA Corporation with the donation of the Tesla K40 GPU used for this research.

\bibliographystyle{IEEEtrans}
\bibliography{egbib}

\end{document}